\documentclass[]{spie}  %>>> use for US letter paper
\usepackage{amsmath,amsfonts,amssymb}
\usepackage{graphicx}
\usepackage[colorlinks=true, allcolors=blue]{hyperref}
\usepackage{textcmds} % for quote
\usepackage{tabularx,booktabs}

\usepackage{array} 
\usepackage{hhline}
\usepackage[caption=false]{subfig}

\title{Data refinement for fully unsupervised visual inspection using pre-trained networks}

\author[1]{Antoine Cordier}
\author[1]{Benjamin Missaoui}
\author[1]{Pierre Gutierrez}

\affil[1]{Scortex, 22 rue Berbier du Mets, Paris, France}

\begin{document} 
\maketitle

\begin{abstract}

\end{abstract}

Anomaly detection has recently seen great progress in the field of visual inspection. More specifically, the use of classical outlier detection techniques on features extracted by deep pre-trained neural networks have been shown to deliver remarkable performances on the MVTec Anomaly Detection (MVTec AD) dataset. However, like most other anomaly detection strategies, these pre-trained methods assume all training data to be normal. As a consequence, they cannot be considered as fully unsupervised. There exists to our knowledge no work studying these pre-trained methods under fully unsupervised setting. In this work, we first assess the robustness of these pre-trained methods to fully unsupervised context, using polluted training sets (i.e. containing defective samples), and show that these methods are more robust to pollution compared to methods such as CutPaste. We then propose SROC, a Simple Refinement strategy for One Class classification. SROC enables to remove most of the polluted images from the training set, and to recover some of the lost AUC. We further show that our simple heuristic competes with, and even outperforms much more complex strategies from the existing literature.


\section{Introduction}
\label{sec:intro}  % \label{} allows reference to this section
Availability of training data holds sway over the applicability of deep learning. In the field of automated quality control, the scarcity of defective parts and the high cost of annotation is heading the industry towards anomaly detection, for which the methods learn to model normality from easily acquired, healthy (i.e. non-defective) images only. However, implicitely assuming the data to be healthy has the consequence that the approach cannot be considered fully unsupervised. In practice, it is always possible for a human operator to inadvertently gather a few defective parts in his predominantly healthy training set, unless he spends time to thoroughly inspect all training parts, which can be a cumbersome task. Indeed, proper labelling may sometimes only be performed by well-trained operators. In this work, we tackle the problem of truly unsupervised anomaly detection for visual inspection, where no labels are available and where the training set may contain defective samples.

We focus on anomaly detection methods that make use of pre-trained networks, since these methods have shown remarkable results in recent works. In particular, we lead comprehensive experiments with four of these methods (KNN\cite{bergman2020deep}, Mahalanobis\cite{rippel2020modeling}, PaDiM\cite{defard2020padim}, and PatchCore\cite{roth2021total}) on the publicly available MVTec AD dataset\cite{bergmann2019mvtec}. Note that the same pre-trained feature extractor (EfficientNetB4\cite{tan2020efficientnet}) is used for all four outlier detectors for fair comparison. Our work brings a twofold contribution:
\begin{enumerate}
  \item We first assess in section \ref{subsec:robustness} the impact of defective training parts on the performances of these detectors, and show that the latter are already quite robust to pollution compared to other methods such as CutPaste\cite{li2021cutpaste}.
  \item We then propose in section \ref{subsec:refinement} a simple yet efficient refinement strategy to remove polluted samples from the training data. Our method is inspired by STOC\cite{yoon2021selftrained}, which applies such a refinement strategy to the CutPaste\cite{li2021cutpaste} anomaly detection method. We apply our own simple strategy to the four aforementioned outlier detection methods. We demonstrate the effectiveness of our method, and show that it can compete with much more complex heuristics in section \ref{subsec:comparison_section}. Interestingly, our refinement strategy is simple enough to be agnostic of the chosen anomaly detection method, and could in theory be used with most anomaly detection methods (e.g. reconstruction or self-supervision based).
\end{enumerate}

Note that we also provide a comprehensive qualitative analysis in order to better understand the robustness of the pre-trained methods to pollution in section \ref{subsec:qualitative}.

\section{Related work}
\label{sec:related_work}  % \label{} allows reference to this section
\subsection{Visual inspection via deep learning}
Visual inspection is a task which can be automated via supervised deep learning \cite{ren2017generic, cordier2021active}, in which a \textbf{convolutional neural network (CNN)} (such as U-Net \cite{ronneberger2015u} or RetinaNet\cite{lin2017focal}) is trained to detect specific defects. Unfortunately, supervised deep learning requires a high volume of annotated images, and cannot handle types of defect which were not seen at training time (like unusual, or theoretical defects). One can either solve the issue in a data-centric manner by generating more data (for example, with the use of synthetic approaches \cite{zambal2019end, gutierrez2021synthetic}), or by shifting the defect detection paradigm to an anomaly detection one, via the use of semi-supervised or unsupervised learning. In this work, we focus on anomaly detection via unsupervised learning.

\subsection{“Unsupervised” learning for anomaly detection}
\textbf{Anomaly detection} has become a very popular field of research lately, especially with regards to computer vision. We focus this review on works which are close to our task, and refer the reader to the Pang et al.\cite{pang2020deep} review from 2020 for a broader look at the topic. We also refer the reader to the review of our previous paper,\cite{gutierrez2021data} which also dealt with anomaly detection for visual inspection using pre-trained networks. We group visual anomaly detection approaches into three main categories: reconstruction of the input, classification of anomalies and modelling of normality.

\textbf{Reconstruction of the input} is most of the time performed using generative convolutional networks, such as auto-encoders (AEs). They have been among the first methods to make use of deep learning for anomaly detection\cite{bergmann2019mvtec,huang2019attribute}. The main objective of these methods consists in learning to reconstruct input images using only normal data for training. At inference time, the network is expected to reconstruct over input anomalies, which allows to compute the anomaly score as a simple L2 distance between the original input image and its reconstruction. This approach has been extended via the use of more appropriate distances for measuring the difference between the input and its reconstruction, such as structural similarity\cite{bergmann2018improving}, as well as the use of more complex generative networks, like variational auto-encoders (VAEs)\cite{baur2021autoencoders} and generative adversarial networks (GANs).\cite{schlegl2017unsupervised}. Since reconstruction based methods have the disadvantage of occasionally reconstructing the anomalies, inpainting techniques have been often used to avoid this phenomenon,\cite{ZAVRTANIK2021107706, nguyen2020unsupervised} sometimes using vision transformers (ViT) to incorporate global context\cite{pirnay2021inpainting}. We refer the reader to Baur et al.\cite{baur2021autoencoders} for an extensive overview of these reconstruction-based methods.

\textbf{Classification of anomalies} is done by leveraging the classification confidences of a CNN, which can be either trained in a direct supervised manner via outlier exposure\cite{hendrycks2018deep}, or in a self-supervised fashion. In the first case, outliers are selected from different distributions, i.e. datasets that are unrelated to the normal training set. Alternatively, outliers can be crafted synthetically from the original training images using local augmentations\cite{daisukelab_2019}. Crafting synthetic outliers can be considered as self-supervised learning, since no manual labelling is required. In CutPaste\cite{li2021cutpaste}, the authors propose to transform any normal training image into an anomalous one by locally incorporating a transformed patch coming from any of the other training images. This allows to learn finer features, which will be more specific to the task-at-hand than outliers coming from a different dataset. This idea has recently been extended by Schlüter et al.\cite{schluter2021selfsupervised}, who train a segmenter network with blended synthetic outliers using soft labels. Note that synthetic outlier exposure can also serve reconstruction-based approaches.\cite{collin2020improved} A slightly different approach is for the CNN to learn an auxiliary task, such as predicting random augmentations applied to the normal input images. In GeoTrans\cite{golan2018deep}, the chosen augmentations are for example rotations and translations. The intuition is that at test time, the augmentations of an out-of-distribution image will not be predicted properly by the model. To some extent, this can be seen as a way to perform synthetic outlier exposure\cite{hendrycks2018deep}. However, it should be noted that the performance of such approaches highly depends on what can be considered as acceptable data augmentations with regards to the dataset of interest, and thus must be designed by hand. This is due to the fact that the trained classifiers are by design discriminative networks, and consequently prone to overfitting to the chosen augmentations. This was pointed out by Zavrtanik et al.\cite{zavrtanik2021draem}, who leverage the predictions of a discriminative network laid on top of a reconstruction one to avoid overfitting, and trained via synthetic outlier exposure. 

\textbf{Modelling normality} for anomaly detection was initially framed as a one-class classification problem by the support vector method (SVM) paper in 1999\cite{scholkopf1999support}. Since then, the use of deep neural networks for learning compact latent distributions has considerably improved these methods, notably Deep SVDD\cite{pmlr-v80-ruff18a} and its pixel-wise extension Patch SVDD \cite{yi2020patch}. Unfortunately, training such networks does not lead to discriminative (or rich) enough features for anomaly detection, and may result in mode collapses under certain conditions. One way to fix this issue is by extending this method to the semi-supervised case, like Deep SAD\cite{ruff2019deep,ruff2020rethinking}, possibly using synthetic outliers \cite{liznerski2020explainable}. Another way to force the network to be descriptive is to start from a pre-trained feature extractor, and to simultaneously learn the compact one-class objective as well as an auxiliary general supervised task on an external dataset, in order to maintain descriptiveness\cite{perera2019learning}. A now common way is to use the features of such a pre-trained feature extractor in combination with traditional anomaly detection methods, as we will see now.

\subsection{Modelling normality with deep pre-trained features}
Building on that idea to balance out descriptiveness and compactness, a recent trend in modelling normality for anomaly detection is to directly take advantage of \textbf{deep feature representations from a frozen pre-trained network}. Using a pre-trained CNN, images are encoded in the embedding space of the network where traditional anomaly detectors can then be used. This guarantees descriptiveness of the features, and avoids mode collapses since the pre-training is performed on a large, generalist dataset such as ImageNet\cite{5206848}. Typical methods for computing the final anomaly score in the embedding space are: distance to the k-nearest neighbors (KNN) \cite{bergman2020deep}, Mahalanobis distance \cite{rippel2020modeling}, and kernel density estimation \cite{erdil2020unsupervised}. Because these methods only provide an anomaly score at the image level and do not necessarily offer anomaly localization, they have been improved in several ways. SPADE\cite{cohen2020sub} and PatchCore\cite{roth2021total} extend the KNN approach by looking at the nearest neighbors locally, in the pixel-embedding space. Correspondingly, PaDiM \cite{defard2020padim} extends the Mahanalobis approach by locally modeling patch-wise gaussian covariances. Finally, in “Deep Feature Reconstruction” (DFR)\cite{yang2020dfr}, Yang et. al. use a reconstruction autoencoder in the pixel-embedding space of the pre-trained feature extractor. Note that the idea of feature reconstruction was first proposed without the need of a frozen pre-trained network, in the “Uninformed Students” paper \cite{bergmann2020uninformed}, where an ensemble of CNNs is trained to mimic a teacher network. The idea has since been extended by using several feature maps from a single network \cite{wang2021student, salehi2020multiresolution}, as well as by blurring the input \cite{choi2019novelty}.

The methods benefiting from frozen pre-trained networks often perform very well because they cannot forget the richness of the pre-trained feature representations, which often happens when fine-tuning such networks on different data due to catastrophic forgetting \cite{goodfellow2013empirical}. By taking advantage of descriptive features, combined with the compactness that the traditional anomaly detectors bring, these methods allow to get state-of-the-art performance on many standard anomaly detection datasets for computer vision, such as MVTec. It is worth noting that techniques such as KNN, Mahalanobis, and PatchCore are also very fast to train and to test  (seconds or minutes) compared to other methods that require an additional CNN training, such as autoencoders or self-supervised networks (hours or days). This makes the said methods prime for small series in quality inspection use cases.

\subsection{Fully unsupervised anomaly detection}
These pre-trained embeddings based methods are also quite robust to the low data regimes, as shown in our previous work.\cite{gutierrez2021data} However, like all of the other presented methods (AEs, self-supervised and one-class approaches), they implicitely assume that the entirety of the training set is normal. In that regard, they cannot be considered as fully unsupervised, since some degree of supervision is required to select the normal training samples. We define \textbf{fully unsupervised learning as a learning which does not require any kind of labelling of the data}. Put differently, a fully unsupervised approach differs from a common unsupervised approach in the fact that it can work with a polluted training set. Examples of fully unsupervised anomaly detection approaches are isolation forest\cite{isolationforest} and local outlier factor\cite{lof}; neither of which relies on deep learning.

This remark is highlighted in STOC\cite{yoon2021selftrained}, which proposes a self-supervised one-class framework for fully unsupervised anomaly detection, by refining the data over multiple iterations of training. Refinement is performed by removing training samples with high anomaly scores, which are computed via the ensembling of classifiers trained on independent splits of the training data. They also assess robustness of the CutPaste\cite{li2021cutpaste} self-supervised method on the MVTec AD dataset\cite{bergmann2019mvtec}, and find that a pollution rate of 10\% leads to a decrease in AUC of around 6 points. To our knowledge, robustness of the pre-trained methods to non-zero anomaly ratios in the training data (i.e. pollution) has not been systematically studied (the original Mahalanobis paper however reports a decrease of AUC of less than 2 points under the same conditions\cite{rippel2020modeling}). In this work, we assess robutness of these methods to pollution, and show that they can be more robust than CutPaste, even without refinement. We also outline a simple yet effective refinement algorithm to recover lost performance. One drawback of data refinement is that it is based on a percentage of data to remove. This percentage is an hyper-parameter that is directly linked to the actual pollution rate, which is not precisely known in practice, but can be estimated based on past observations.

\section{Materials and methods}
\label{sec:methods}  % \label{} allows reference to this section

\subsection{Pre-trained networks for anomaly detection}
\label{subsec:pre-trained}
Anomaly detection with pre-trained networks attempts to model normality solely from healthy samples. In this work, our goal is to assess and improve the performances of these methods when the training set contains defective samples. We will be comparing KNN\cite{bergman2020deep}, Mahalanobis\cite{rippel2020modeling}, PaDiM\cite{defard2020padim} and the recently introduced PatchCore \cite{roth2021total} in the following experiments. As in-depth explanations of KNN, Mahalanobis and PaDiM are already available in our previous work \cite{gutierrez2021data}, we refer to the latter for more details on these methods.

\subsubsection{KNN}
\label{subsubsec:spade}
Introduced by Bergman et al., the KNN \cite{bergman2020deep} uses a frozen feature extractor $f$ to embed the training set into a lower dimensional space. Global Average Pooling (GAP) is then applied to these feature maps to obtain a final vector representation for each image. At test time, the anomaly score of a new image $y$ is calculated by taking the average euclidean distance of its associated vector representation $f_y$ to its $k$ nearest neighbours in the training set $N_k(f_y)$:
\begin{equation}
d(y) = \frac{1}{k}\sum_{f \in N_k(f_y)} ||f-f_y||^2
\end{equation}
In our experiments, we implement SPADE\cite{cohen2020sub}, a pixel-wise extension of the KNN which allows localization of the anomalies. SPADE uses the exact same image-wise anomaly score, but also gives an anomaly score for each patch $(i,j)$ by calculating the average euclidean distance between the test feature vector $f_{y_{i,j}}$ and the feature vectors retrieved by the KNN $f_{x^k_{i,j}}$. Also, SPADE applies GAP to multiple feature levels and concatenate the results to obtain a vector representation of the image with both coarse and fine-grained features. The authors, as well as later work \cite{rippel2020modeling}, show the effectiveness of using multiple feature levels. We set the number of nearest neighbors
$k$ to 5.

\subsubsection{Mahalanobis}
\label{subsection:mahalanobis}
Rippel et al.\cite{rippel2020modeling} propose to model the distribution of training embeddings with a Multivariate Gaussian distribution (MVG) and use the Mahalanobis distance as the anomaly score. More specifically, they estimate the mean and covariance of the training embeddings at different feature levels. Then, they compute the distance to the distribution of the test embeddings $f_y$ at each of these levels:
\begin{equation}
d(y) = \sqrt{(f_y-\mu)^TS^{-1}(f_y-\mu)}
\end{equation}
The distances are summed up across all feature levels to give the final anomaly score. As our goal is to find a strategy to refine the polluted training set (which will inevitably reduce the number of training samples), we use Ledoit's shrinkage\cite{LEDOIT2004365} as the covariance estimator, which is known to perform better when the number of datapoints becomes much lower than the number of features\cite{gutierrez2021data}. In order to get localization for this method, we apply Grad-CAM\cite{gradcam2019}.

\subsubsection{PaDiM}
Proposed by Defard et al., PaDiM\cite{defard2020padim} can be seen as a pixel-wise extension of the Mahalanobis approach. The features from the different levels are aligned and concatenated, before fitting a MVG at every spatial location of the training images. At test time, PaDiM calculates the Mahalanobis distance of every patch $y_{i,j}$ of the test image $y$ to the corresponding MVG. The image-wise anomaly score is the maximum of the scores at every location:
\begin{equation}
d(y) = \max_{i,j} \sqrt{(f_{y_{i,j}}-\mu_{i,j})^TS_{i,j}^{-1}(f_{y_{i,j}}-\mu_{i,j})}
\end{equation}
Like for Mahalanobis (see \ref{subsection:mahalanobis}), we also use Ledoit's shrinkage as our covariance estimator in order to get a more robust estimation when the number of datapoints becomes much lower than the number of features.

\subsubsection{PatchCore}
Recently introduced by Roth et al, PatchCore\cite{roth2021total} can be seen as another pixel-wise extension of the KNN. PatchCore computes and stores a vector representation $f_{x_{i,j}^n}$ for all patches $(i,j)$ of all training images $x^n, n \in \{1,...,N\}$. At test time, it computes a representation $f_{y_{i,j}}$ for every patch location $(i,j)$ of the test image $y$, and attempts to retrieve the $k$ nearest embeddings for every location. The patch-level anomaly score is given by the average euclidean distance from $f_{y_{i,j}}$ to its $k$ nearest embeddings $N_k(f_{y_{i,j}})$. The image-level anomaly score is the maximum score across all spatial locations:
\begin{equation}
d(y) = \max_{i,j} \frac{1}{k}\sum_{f_{i,j} \in N_k(f_{y_{i,j}})} ||f_{i,j}-f_{y_{i,j}}||^2
\end{equation}
In this approach, the $k$ nearest embeddings of $f_{y_{i,j}}$ can come from any training image, at any location. This gives PatchCore a lot more flexibility  with regards to the data distribution (multimodality), and allows for precise defect segmentation.  In order to alleviate the time and memory constraints of the nearest neighours search, the authors propose to use coreset selection\cite{Agarwal05geometricapproximation} to find a representative subsample of the training embeddings, and perform nearest neighbors retrieval in this reduced space. In our implementation of PatchCore, we do not use coreset but IVFFlat (Inverted File Index Flat)\cite{5432202} as an acceleration scheme, as it has showed better results in our experimentations. We use the faiss library\cite{JDH17} for IVFFlat implementation, and we set the number of nearest neighbors $k$ to 5. Note the two main differences with SPADE: here the image-wise anomaly score is the score of the highest patch instead of being the score of an image-wise KNN, and the nearest neighbors of a test patch $y_{i,j}$ can come from any image in the training set at any location, instead of being limited to patches retrieved by the image-wise KNN.

\subsection{SROC, a simple heuristic for data refinement}
\label{subsec:methodology}

% SROC figure
\begin{figure}
  \centering
  \includegraphics[width = 380pt]{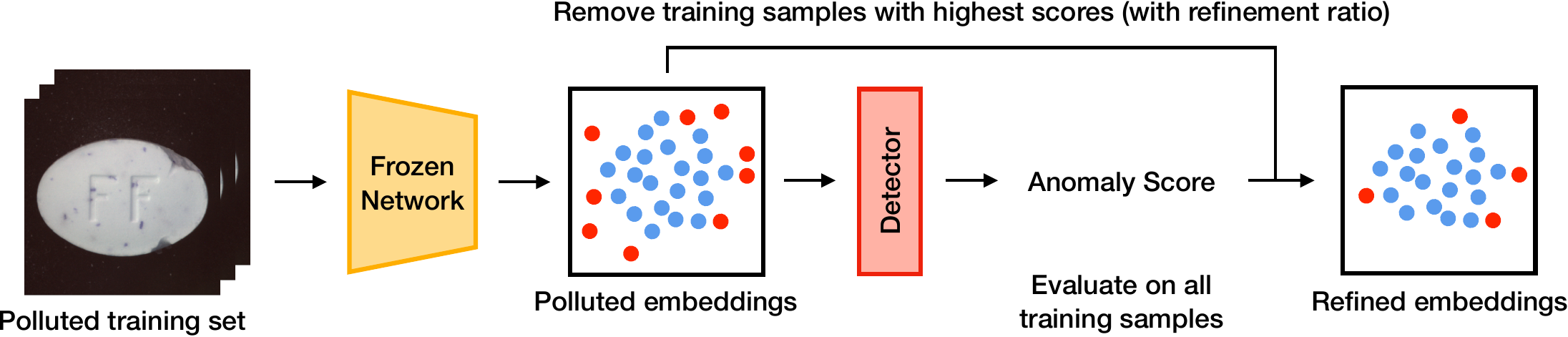}
  \caption{Overview of SROC. A frozen feature extractor is used to embed the polluted training set, and a one-class classifier (or detector) is fitted on these embeddings. The detector then predicts a score for each training samples, and the samples with the highest scores are removed.}
  \label{fig:naive_figure}
\end{figure}

\subsubsection{Methodology}

We propose SROC, a \textbf{S}imple \textbf{R}efinement strategy for \textbf{O}ne \textbf{C}lass anomaly detection. The goal of our method is to leverage both the high proportion of healthy samples in the training data and the good generalization capabilities of the traditional one-class classifiers.

Here are the steps we follow. First, a pre-trained feature extractor is used to extract a rich and meaningful representation of the polluted training set. Then, an anomaly detector from the ones detailed in \ref{subsec:pre-trained} is fitted on the polluted embeddings. We evaluate the detector on the same training embeddings, and we remove samples with the greatest anomaly scores. Finally, another detector can then be fitted on the refined embeddings to reach better performances. Figure \ref{fig:naive_figure} provides a visual representation of the proposed method. 

This strategy is based on the assumption that most of the training data is healthy, because manufacturers optimize their production to yield as few defective parts as possible. By leveraging this with the generalization capabilities of the detectors, we aim at constructing a representation of normality that is good enough for training defective samples to have a higher score. In this paper, we validate the effectiveness of SROC with the 4 state-of-the-art detectors presented in section \ref{subsec:pre-trained}, but our heuristic has the advantage to be compatible with virtually any one-class classifier.

\subsubsection{Hyperparameters}

SROC is designed to be simple and easy to implement in a production setting, where precise hyperparameter tuning for each part reference is not always feasible. The only parameter of the method is the percentage of samples to remove, which we call the refinement ratio. Ideally, the refinement ratio should be close to the actual pollution rate, but can also be higher, as already shown in STOC\cite{yoon2021selftrained}.

\section{Experiments}
\label{sec:experiments}

\subsection{Dataset and metrics}
\label{subsec:datasets}

\begin{figure}[hb!]
  \centering
  \includegraphics[width = 280pt]{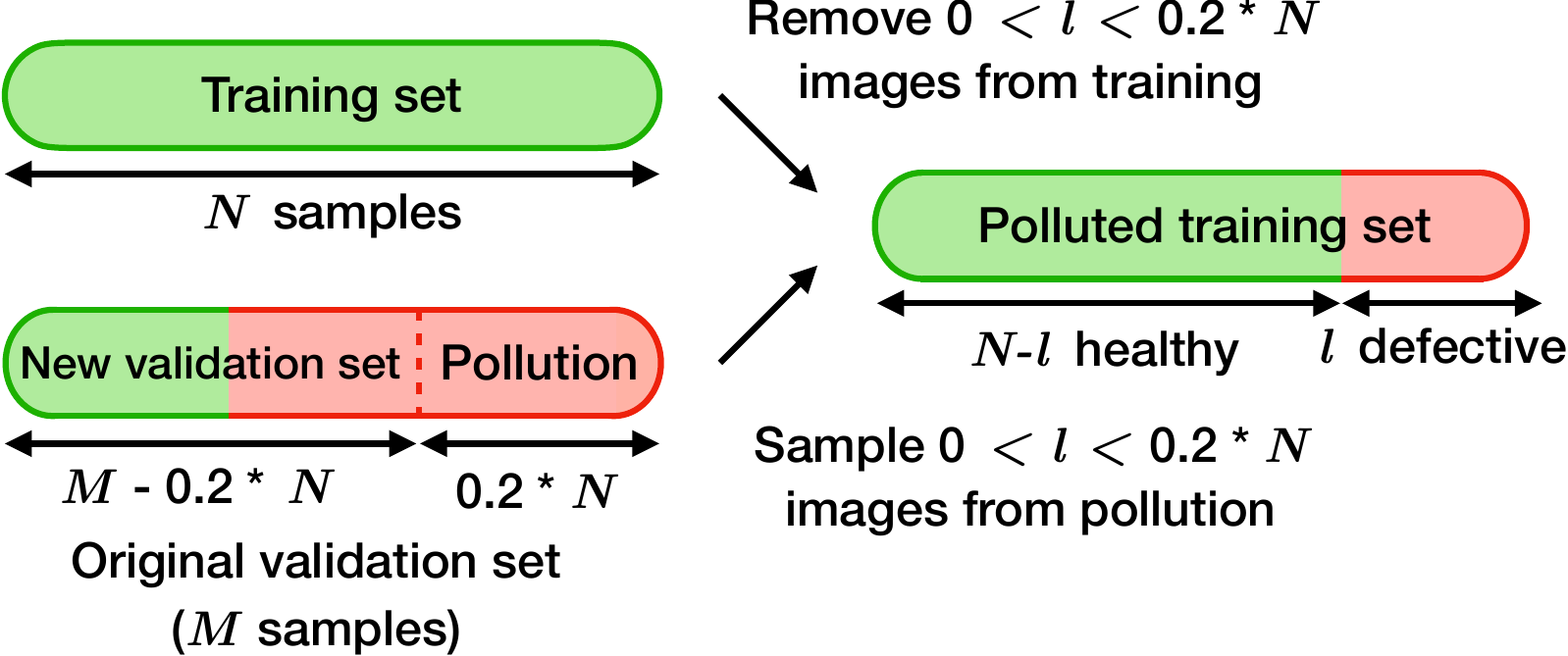}
  \caption{Creation of the MVTec polluted training sets. A set of defective samples is extracted from the original validation set, and gradually injected in the training set in place of healthy samples for pollution purposes. The resulting training set has the same size as the original.}
  \label{fig:polluted_datasets_creation}
\end{figure}

We conduct the experiments on MVTec AD\cite{bergmann2019mvtec}, a dataset for real-world industrial quality inspection comprised solely of healthy samples for training and of both healthy and defective samples for validation. Images are divided into 15 classes or categories: some being textures (e.g. \textit{carpet}, \textit{grid}, \textit{leather}) and others being objects (e.g. \textit{bottle}, \textit{cable}, \textit{capsule}).

Since no defects are available in the original training sets, we create polluted training sets in a similar fashion as in the STOC paper\cite{yoon2021selftrained}. For each class, we first sample a “pollution set” from the original validation images that is 20\% the size of the training set $N$. This pollution set contains only defective samples. Note that we ensure that the defects sampled in the pollution set follow the same distribution as the defects in the original validation set (i.e. there is the same proportion of each defect type in both). Samples used in the pollution set are excluded from the validation set. Then, we remove $l$ random images from the training set and replace them with random images from the pollution set. $l$ varies from 0 to $0.2 * N$, enabling us to have training sets with 0 to 20\% pollution. Figure \ref{fig:polluted_datasets_creation} summarizes the creation of the polluted training sets. Note that we exclude \textit{hazelnut} and \textit{transistor} categories from our experiments, since their original validation sets do not contain enough defective samples for polluting the training set while keeping defective samples in the validation set. While this method changes the size of the validation set, it has the advantage of preserving the size of the training set. This is to allow measuring the impact of pollution independently of the number of training samples.

We first assess the robustness of the detectors by measuring their AUC, AU-IoU (area under the “IoU as a function of false positive rate" curve) and AU-PRO (area under the “PRO as a function of false positive rate" curve) on the validation set as a function of the pollution ratio. The PRO score (Per-Region Overlap) has been used in previous works\cite{bergmann2019mvtec, bergmann2020uninformed} as an alternative to the pixel-wise AUC and IoU metrics, which tend to be biased towards larger anomalous regions. Here, we measure the AU-IoU and AU-PRO with up to a 30\% false positive rate (FPR) in order for our results to be comparable with previous works \cite{bergmann2019mvtec, bergmann2020uninformed}. Note that this prevents in practice the metrics from saturating between 30\% and 100\% FPR.

We then evaluate the efficiency of SROC as a data refinement strategy by fixing the pollution ratio in the training set to 20\%, and varying the refinement ratio. We compute the same metrics, as well as the precision, recall and F1 score of SROC on the training set. For example, the recall score of SROC is defined as the number of defective samples removed, divided by the total number of defective samples to remove in the training set.

\subsection{Implementation details}
\label{subsec:implementation_details}

In order to be fair, the same feature extractor is used for all methods, namely EfficientNetB4\cite{tan2020efficientnet}, which showed the best results in previous works\cite{rippel2020modeling}. We use the outputs of the blocks 4, 6 and 7, as in PaDiM's original paper\cite{defard2020padim}. These choices may be slightly detrimental to some methods and can explain the performance differences with the ones reported by the original authors.

We implement all methods using Tensorflow / Keras for embedding the images with EfficientNetB4, as well as Scikit-Learn and faiss\cite{JDH17} for the traditional anomaly detection techniques. The hyperparameters chosen for each detector are presented in section \ref{subsec:pre-trained}. Since we are using EfficientNetB4, we resize all MVTec images to 380 x 380 pixels (the image size with which the network was originally pre-trained). Note that we do not perform center cropping, as it is common to do. This may also explain differences in performances.

We conduct the experiments with 5 different random seeds to provide both mean results and standard deviations. The seed controls which healthy samples are removed from the training set, and which defective samples are added in place. Note that the 5 seeds choice is the same for all pollution and refinement ratios.

\subsection{Assessing pollution robustness}
\label{subsec:robustness}
\begin{figure}[h!]
  \centering
  \includegraphics[width = 300pt]{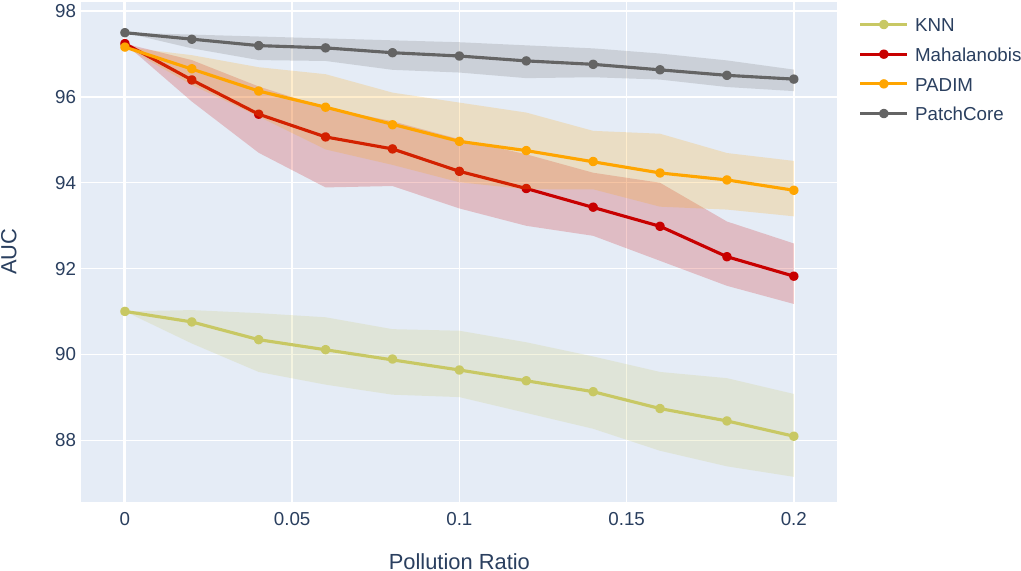}
  \caption{Average AUC across all MVTec AD categories as a function of the pollution ratio, for each detector. Categories \textit{transistor} and \textit{hazelnut} were removed because of their low amount of available defective samples.}
  \label{fig:robustness_AUC}
\end{figure}

We start by assessing the robustness of each anomaly detection method to increasingly polluted training sets. To that end, we fit the different detectors on training sets with pollution ratios varying between 0 and 20\%, and evaluate them on the validation sets defined in \ref{subsec:datasets}. Table \ref{tab:robustness_recap_table} compares the performances of all detectors at exactly 0 and 20\% pollution ratios. Figure \ref{fig:robustness_AUC} shows the evolution of the AUC as the pollution ratio increases. Note that the AU-IoU and AU-PRO as functions of the pollution ratio are available in Appendix \ref{appendix:add-results-robust}.

\newcolumntype{Y}{>{\centering\arraybackslash}X}

  \begin{table}[ht!]
    \caption{Average performances of each detector across all MVTec AD categories, when the training set is polluted at 0 and 20\%. Categories \textit{transistor} and \textit{hazelnut} were removed because of their low amount of available defective samples.}
    \label{tab:robustness_recap_table}
    %\begin{tabular}{|>{\centering\arraybackslash}m{1.8cm}|c|c|c|c|c|c|c|c|} % Set the width of the first column to prevent the table from being too large
    \smallskip
    \begin{tabularx}{\textwidth}{l*{9}{Y}}
     \toprule
     &\multicolumn{2}{c}{KNN}
     &\multicolumn{2}{c}{Mahalanobis}
     &\multicolumn{2}{c}{PaDiM}
     &\multicolumn{2}{c}{PatchCore} \\
     \cmidrule(lr){2-3}\cmidrule(lr){4-5}\cmidrule(lr){6-7}\cmidrule(lr){8-9}
     %\hline
     %\midrule
     Pollution 
     & 0\% & 20\% 
     & 0\% & 20\%
     & 0\% & 20\%
     & 0\%& 20\%\\
     %\hline
     \midrule
     carpet
     & 97.8 & 93.3±0.5
     & 100 & 99.5±0.1 
     & 100 & 100±0.0 
     & 100 & 100±0.0\\
     %\hline
    tile
     & 99.5 & 99.5±0.2
     & 100 & 99.8±0.1 
     & 99.7 & 99.2±0.2 
     & 99.9 & 99.5±0.2\\
     %\hline
    leather
     & 99.9 & 98.9±0.1 
     & 100 & 95.3±0.4 
     & 100 & 100±0.0 
     & 100 & 100±0.0\\
     %\hline
     grid
     & 67.9 & 60.0±3.1
     & 97.6 & 82.9±1.9
     & 100 & 100±0.0
     & 100 & 100±0.0\\
     %\hline
     wood
     & 87.6 & 79.7±1.7
     & 95.2 & 85.2±1.0
     & 100 & 99.0±0.0
     & 100 & 100±0.0\\
     %\hline
     capsule
     & 92.7 & 91.9±1.6
     & 96.7 & 91.4±1.2
     & 95.5 & 81.7±2.8
     & 98.2 & 95.6±1.0\\
     %\hline
     cable
     & 95.3 & 94.6±0.2
     & 97.9 & 94.2±0.4
     & 97.7 & 94.2±0.5
     & 97.4 & 94.2±0.5\\
     %\hline
     pill
     & 80.4 & 78.4±0.9 
     & 88.8 & 84.0±0.4 
     & 94.6 & 88.5±0.7 
     & 93.4 & 90.4±0.4\\
     %\hline
     metal nut
     & 88.6 & 84.5±1.5 
     & 97.5 & 93.9±0.1 
     & 95.5 & 88.4±0.7
     & 100 & 99.3±0.2\\
     %\hline
     toothbrush
     & 93.5 & 90.6±1.9 
     & 96.8 & 95.0±1.3
     & 88.0 & 86.4±1.6
     & 89.4 & 89.4±0.0\\
     %\hline
     screw
     & 85.7 & 81.3±2.7 
     & 95.5 & 81.0±2.6
     & 96.2 & 93.0±1.2
     & 93.6 & 91.8±1.4\\
     %\hline
     zipper
     & 95.3 & 93.8±0.3
     & 98.1 & 93.1±1.0
     & 96.3 & 90.8±1.2
     & 96.3 & 94.0±0.1\\
     %\hline
     bottle
     & 98.8 & 98.7±0.3
     & 100 & 98.4±0.2
     & 99.5 & 98.5±0.9
     & 99.3 & 99.0±0.0
     \\
     %\hline
     \textbf{Mean AUC} & 91.0 & 88.1±1.0 & 97.2 & 91.8±0.8 & 97.2 & 93.8±0.7 & 97.5 & 96.4±0.3\\
     %\hhline{|=|=|=|=|=|=|=|=|=|}
     \midrule
     \textbf{Mean AU-IoU}
     & 0.164 & 0.135 & 0.142 & 0.121 
     & 0.147 & 0.128 & 0.167 & 0.141\\
     %\hline
     \midrule
     \textbf{Mean AU-PRO}
     & 0.909 & 0.902 & 0.810 & 0.804
     & 0.877 & 0.858 & 0.914 & 0.908\\
    %\hline
    %\end{tabular}
    \bottomrule
    \end{tabularx}
   \end{table}
   
Results show that detectors leveraging pre-trained feature extractors are already quite robust to polluted training set, even at 20\%. Mahalanobis and PaDiM are the most impacted out of the four (-5.4 and -3.4 AUC points, respectively), and PatchCore and KNN are the most robust (-1.1 and -2.9 AUC points, respectively). In comparison, the authors of STOC report that CutPaste\cite{li2021cutpaste} trained on a dataset with a 10\% pollution ratio looses 5.97 AUC points \cite{yoon2021selftrained} on MVTec AD. As a consequence, anomaly detection methods leveraging a pre-trained feature extractor appear as more robust to pollution than purely self-supervised methods. We explan these discrepancies by the use of a frozen feature extractor. The latter preserves the richness and descriptiveness of the ImageNet features. The pollution only impacts the shallow anomaly detector on top. Self-supervised methods, on the other hand, fit the feature extractor direcly onto the polluted images, thus impacting the network representation. It therefore becomes much harder for the detector on top to discriminate between the healthy and the defective images.

Note however that our results are not directly comparable to the ones reported for CutPaste, mainly for two reasons. First, even though the polluted training set creation approach is the same as the one reported for CutPaste, the latter only pollute at a 10\% ratio at most, making the final validation sets different from ours (not to mention that the sampling of the defective parts from the original validation set is done at random, which could lead to different results even if the maximum pollution ratio was the same). Second, we had to exclude \textit{transistor} and \textit{hazelnut} categories because of their lack of defective validation samples.

Interestingly, nearest neighbors based detectors (KNN, PatchCore) appear as more robust than the ones relying on Mahalanobis distance (Mahalanobis, PaDiM). In section \ref{subsec:qualitative}, we conduct an in-depth qualitative analysis in order to understand why.

\subsection{Data refinement}
\label{subsec:refinement}

\begin{figure}[hb!]
  \centering
  \includegraphics[width = 300pt]{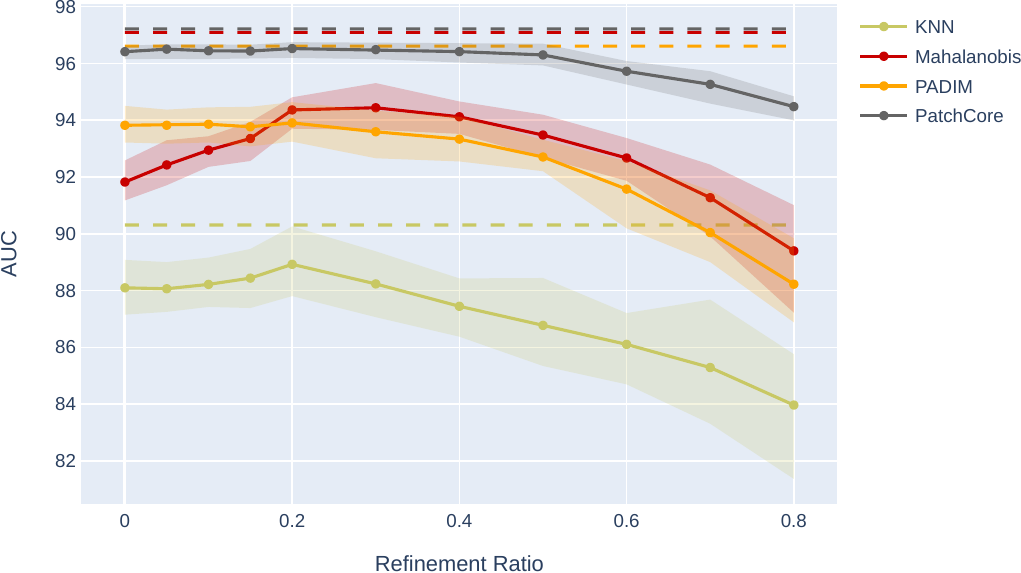}
  \caption{Average AUC for all MVTec classes as a function of the refinement ratio, for training sets polluted at a 20\% ratio. Dashed lines represent maximum reachable AUC for a perfect strategy which would only removes the defective samples (average performances of each detector when fitted on 80\% of the original healthy training set).}
  \label{fig:naive_AUC}
\end{figure}

Despite the relative robustness of the pre-trained methods, their loss of performance could deter their use in a fully unsupervised setting. In this section, we show the effectiveness of our method for data refinement to alleviate this loss. We pollute all training sets with 20\% of defective samples and refine them with SROC. We compare the performances of each detector when removing from 0\% to 80\% of the top predictions. Figure \ref{fig:naive_AUC} and Table \ref{tab:naive_recap_table} give an overview of the performances of the proposed method. 

For all detectors, the highest F1-score is obtained when the refinement ratio is approximately equal to the pollution ratio (here 20\%). At this value, SROC systematically removes more than 70\% of the defective samples from the training set, enabling Mahalanobis and KNN to gain back 2.6 and 0.8 AUC points respectively when fitted on these refined embeddings. After this, the AUC gradually decreases because we are removing too many images. PatchCore does not really improve, but it was also the least impacted by pollution (see section \ref{subsec:robustness}). More surpringly however, PaDiM does not improve. Indeed, it is unclear why SROC does not manage to remove as many defective samples for that method, as shown by the lower precision and recall. Additionally, we have shown in previous works that PaDiM is the most data hungry method out of the four tested for MVTec object categories. Thus, PaDiM is also expected to be more sensible to data refinement than the other methods.

Because SROC works best with Mahalanobis (highest recall and precision), we propose to use SROC+Mahalanobis to refine the embeddings, and to fit KNN, PaDiM, and PatchCore on these refined embeddings. We show in Appendix \ref{appendix:maha-refine} that this enables to gain back 1.4 AUC points for PaDiM.

% ===== TRANSPOSED TABLE ===== 
  \begin{table}[h!]
    \caption{\label{tab:naive_recap_table}Average performances of SROC applied with 0\%, 20\% and 40\% refinement ratios, for training sets polluted at a 20\% ratio. Categories \textit{transistor} and \textit{hazelnut} were removed because of their low amount of available defective samples. Note that we do not display standard deviations here for clarity purposes.}
    \smallskip
    \begin{tabularx}{\textwidth}{l*{13}{Y}}
    \toprule
     &\multicolumn{3}{c}{KNN}
     &\multicolumn{3}{c}{Mahalanobis}
     &\multicolumn{3}{c}{PaDiM}
     &\multicolumn{3}{c}{PatchCore}\\
     \cmidrule(lr){2-4}\cmidrule(lr){5-7}\cmidrule(lr){8-10}\cmidrule(lr){11-13}
     Refinement & 0\% & 20\% & 40\% & 0\% & 20\% & 40\% & 0\% & 20\% & 40\% & 0\% & 20\% & 40\%\\
     %\hline
     \midrule
     carpet
     & 93.3 & 97.8 & 97.8 & 99.5 & 99.8 & 100 & 100 & 100 & 100 & 100 & 100 & 100\\
    % \hline
    tile
     & 99.5 & 99.5 & 99.2 & 99.8 & 100 & 100 & 99.2 & 98.9 & 98.9 & 99.5 & 99.7 & 99.5\\
     %\hline
    leather
     & 98.9 & 99.9 & 99.9 & 95.3 & 100 & 100 & 100 & 100 & 100 & 100 & 100 & 100\\
     %\hline
     grid
     & 60.0 & 55.5 & 48.8 & 82.9 & 85.5 & 84.0 & 100 & 100 & 100 & 100 & 100 & 100\\
     %\hline
     wood
     & 79.7 & 86.8 & 86.4 & 85.2 & 94.4 & 95.9 & 99.0 & 100 & 100 & 100 & 100 & 100\\
     %\hline
     capsule
     & 91.9 & 91.0 & 89.9 & 91.4 & 93.1 & 93.5 & 81.7 & 83.1 & 82.8 & 95.6 & 94.8 & 93.4\\
     %\hline
     cable
     & 94.6 & 94.8 & 95.3 & 94.2 & 96.6 & 96.5 & 94.2 & 95.7 & 96.2 & 94.2 & 95.8 & 95.3\\
     %\hline
     pill
     & 78.4 & 78.7 & 77.6 & 84.0 & 86.6 & 85.5 & 88.5 & 87.6 & 83.9 & 90.4 & 90.3 & 90.4\\
     %\hline
     metal nut
     & 84.5 & 86.4 & 81.4 & 93.9 & 95.6 & 93.2 & 88.4 & 87.4 & 86.7 & 99.3 & 100 & 100\\
     %\hline
     toothbrush
     & 90.6 & 90.6 & 89.4 & 95.0 & 96.2 & 94.6 & 86.4 & 85.8 & 86.8 & 89.4 & 89.3 & 89.3\\
     %\hline
     screw
     & 81.3 & 81.6 & 78.0 & 81.0 & 83.2 & 84.0 & 93.0 & 90.3 & 84.2 & 91.8 & 91.6 & 90.5\\
     %\hline
     zipper
     & 93.8 & 94.7 & 94.6 & 93.1 & 96.2 & 96.8 & 90.8 & 94.2 & 94.3 & 94.0 & 94.3 & 95.6\\
     %\hline
     bottle
     & 98.7 & 98.7 & 98.5 & 98.4 & 99.4 & 99.5 & 98.5 & 97.8 & 99.8 & 99.0 & 99.0 & 99.3\\
     %\hline
     \textbf{Mean AUC}
     & 88.1 & \textbf{88.9} & 87.4 & 91.8 & \textbf{94.4} & 94.1 & 93.8 & \textbf{93.9} & 93.3 & 96.4 & \textbf{96.5} & 96.4\\
     %\hhline{|=|=|=|=|=|=|=|=|=|=|=|=|=|}
     \midrule
     \textbf{Mean AU-IoU}
     & 0.135 & 0.166 & 0.165 & 0.121 & 0.141 & 0.141 & 0.128 & 0.135 & 0.143 & 0.141 & 0.151 & 0.162\\
     %\hline
     \midrule
     \textbf{Mean AU-PRO}
     & 0.902 & 0.906 & 0.906 & 0.804 & 0.794 & 0.797 & 0.858 & 0.862 & 0.865 & 0.908 & 0.907 & 0.907\\
     \midrule
     %\hhline{|=|=|=|=|=|=|=|=|=|=|=|=|=|}
     \textbf{Refin. Precision}
     & - & 78.6 & 45.5 & - & 81.1 & 47.1 & - & 71.5 & 46.4 & - & 75.2 & 46.3\\
     %\hline
     \textbf{Refin. Recall}
     & - & 78.6 & 91.2 & - & 81.1 & 94.3 & - & 71.5 & 93.0 & - & 75.2 & 92.7\\
     %\hline
     \textbf{Refin. F1}
     & - & \textbf{78.6} & 60.7 & - & \textbf{81.1} & 62.8 & - & \textbf{71.5} & 61.9 & - & \textbf{75.2} & 61.7\\
     \bottomrule
    \end{tabularx}
  \end{table}

\subsection{Comparison with other data refinement strategies}
\label{subsec:comparison_section}

In this section, we provide a comprehensive comparison between our refinement strategy and several others, from the literature and from our own experiments.

\textbf{Random}: The simplest heuristic for data refinement is to remove random training samples, and hope to remove the defective ones. This idea is motivated by the good performances obtained by pre-trained detectors in the low data regime.\cite{gutierrez2021data} This will serve as baseline to evaluate the need for a slightly more elaborate strategy like ours.

\textbf{Cross-Validation}: One criticism that could be made about our method is that evaluating a detector on the training set is prone to overfitting. The detector could learn the defects of the training set and therefore give them a low score at test time. We thus compare our method with a cross-validation strategy, where we split the training embeddings into $s$ non-overlapping splits, we fit the detector on $s-1$ splits and evaluate on the remaining split. By iterating over all splits, every training sample gets an anomaly score. Then, we remove the images with the highest scores.

\begin{figure}
\centering
\subfloat[KNN]{\includegraphics[width=0.5\linewidth]{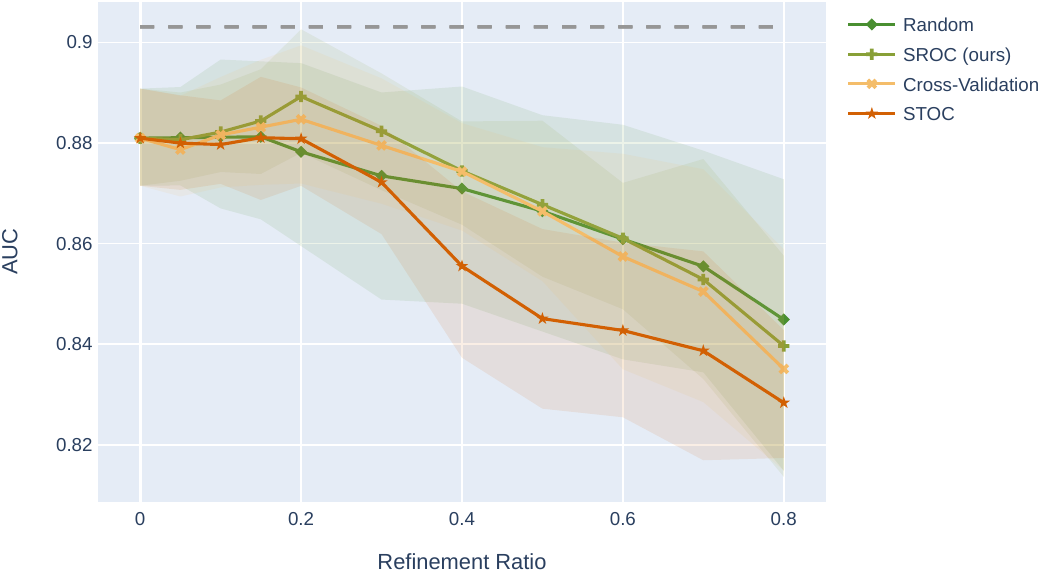}}
\subfloat[Mahalanobis]{\includegraphics[width=0.5\linewidth]{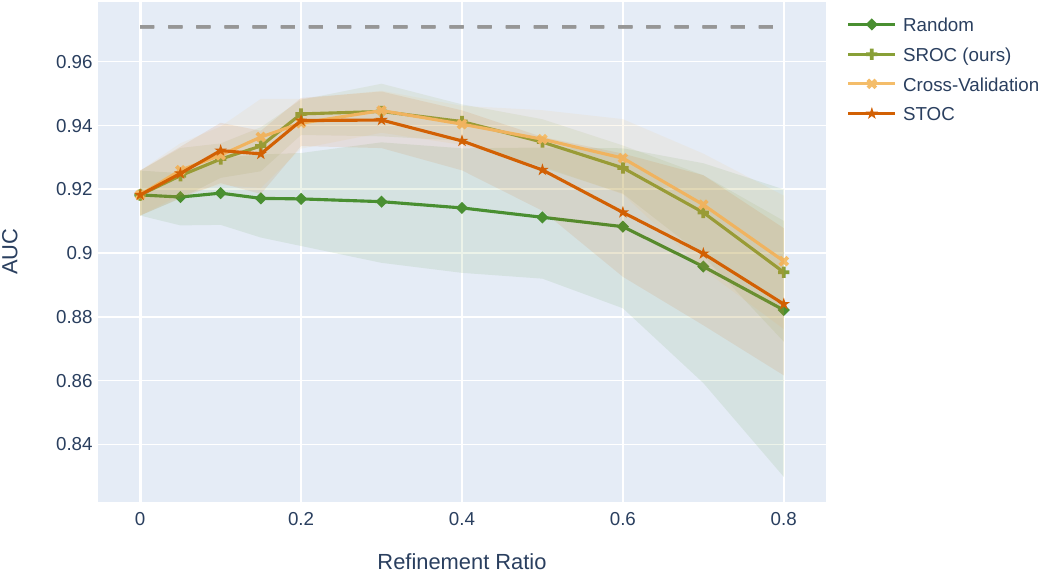}}\\
\subfloat[PaDiM]{\includegraphics[width=0.5\linewidth]{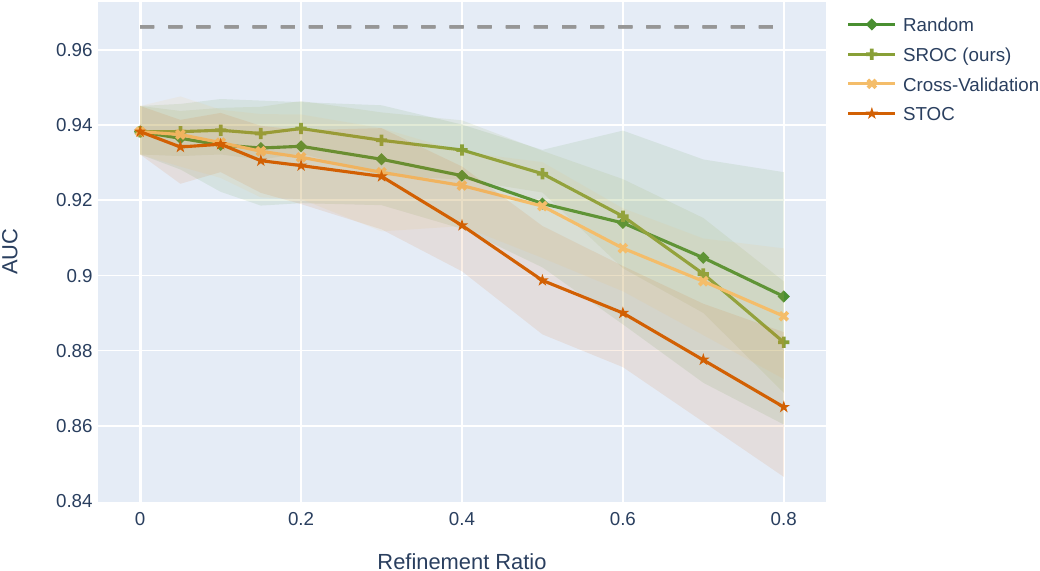}}
\subfloat[PatchCore]{\includegraphics[width=0.5\linewidth]{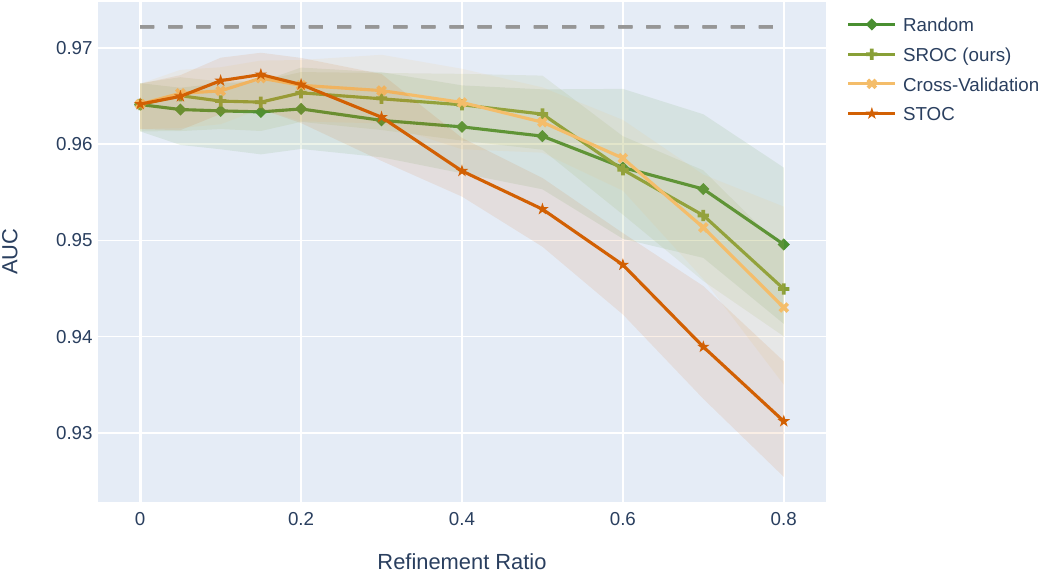}}
\smallskip
\caption{AUC as a function of the refinement ratio for training sets polluted at a 20\% ratio, for all refinement strategies. Results are plot per detector. Dashed lines are the average performances of each detector when fitted on 80\% of the original healthy training set (maximum reachable if the strategy only removes defective samples).  Our simple heuristic competes with much more complex and costly strategies like STOC.}
\label{fig:strategies_comparison}
\end{figure}

\textbf{STOC}\cite{yoon2021selftrained}: While mainly designed for self-supervised methods, STOC could also be used for pre-trained ones. Here, the training set is divided into $s$ splits, a detector is fitted on each of them and evaluated on the whole training set. Each training image is thus given $s$ scores, and we average them to get the final one. The images with the highest scores are removed. This strategy has the advantage of ensembling the decisions of multiple detectors, but each of them is fitted on a fraction of the training data. Note that contrarily to the original paper, we do not apply this strategy iteratively as our feature extractor is frozen. We set the number of splits to $s=5$ for Cross-validation and STOC (as in STOC's original paper\cite{yoon2021selftrained}). Figure \ref{fig:strategies_comparison} compares the different strategies for all detectors. 

Our simple heuristic manages to compete with, and most often outperform both Cross-Validation and STOC. We explain the performance discrepancies by the size of the training set. Our method, SROC, is fitted on the entire training set, while Cross-Validation is trained with the images of $s-1$ splits (here 4/5 of the training data). For STOC, the training size is even smaller as each detector is fitted on a single split $s$ (here 1/5 of the training set). Using all of the available examples for training enables our method to build a more robust and reliable representation of normality, making the polluted samples stand out more from the rest.
SROC is also much more computationally efficient than the others. It only fits a single detector on the polluted embeddings. The latter two, on the other hand fit as many detectors as the number of splits, thus multiplying both the training and the inference time. The simplicity, performance and efficiency of SROC make it a method of choice for a deployment in a production setting.

\subsection{Qualitative analysis, robustness of pre-trained method}
\label{subsec:qualitative}

Finally, in this last section we aim at providing more intuition on the results of section \ref{subsec:robustness}. In particular, our goal is to explain why KNN and PatchCore are less impacted by pollution than Mahalanobis and PaDiM.

\begin{figure}
\centering
\subfloat[carpet]{\includegraphics[width=.4\linewidth]{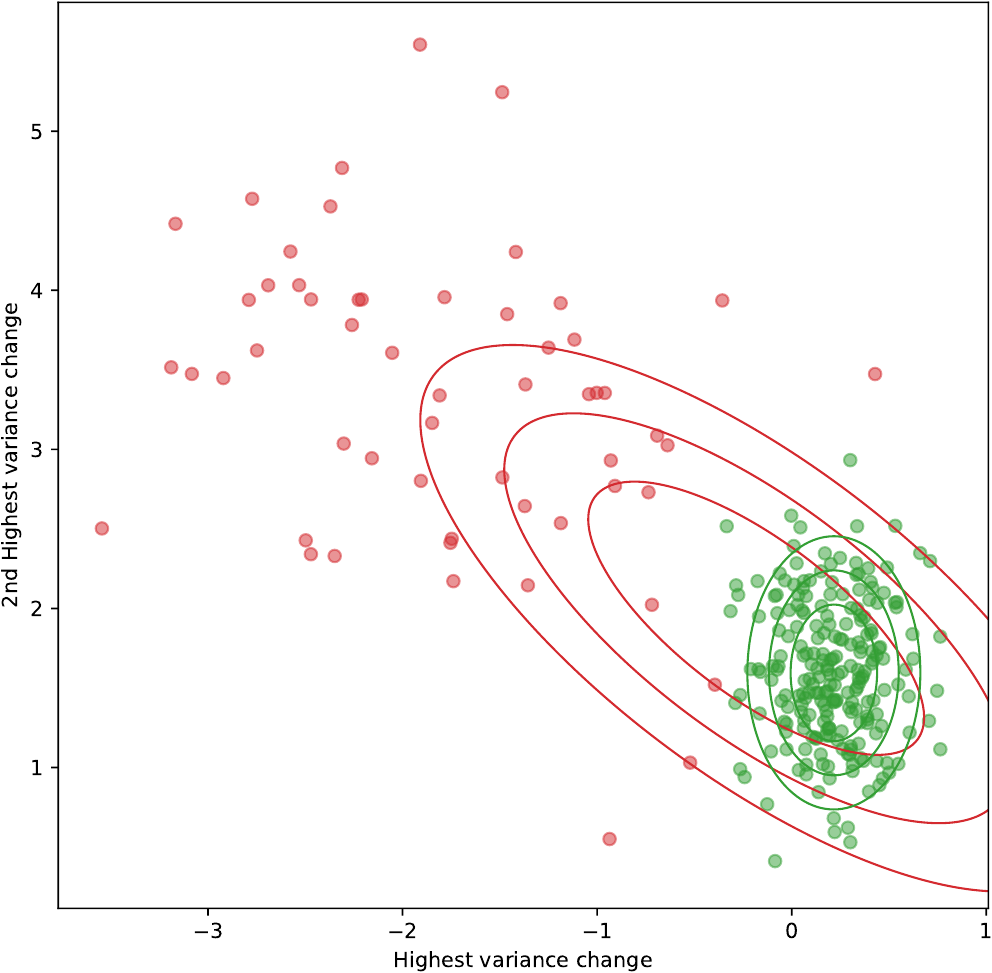}}\quad
\subfloat[bottle]{\includegraphics[width=.4\linewidth]{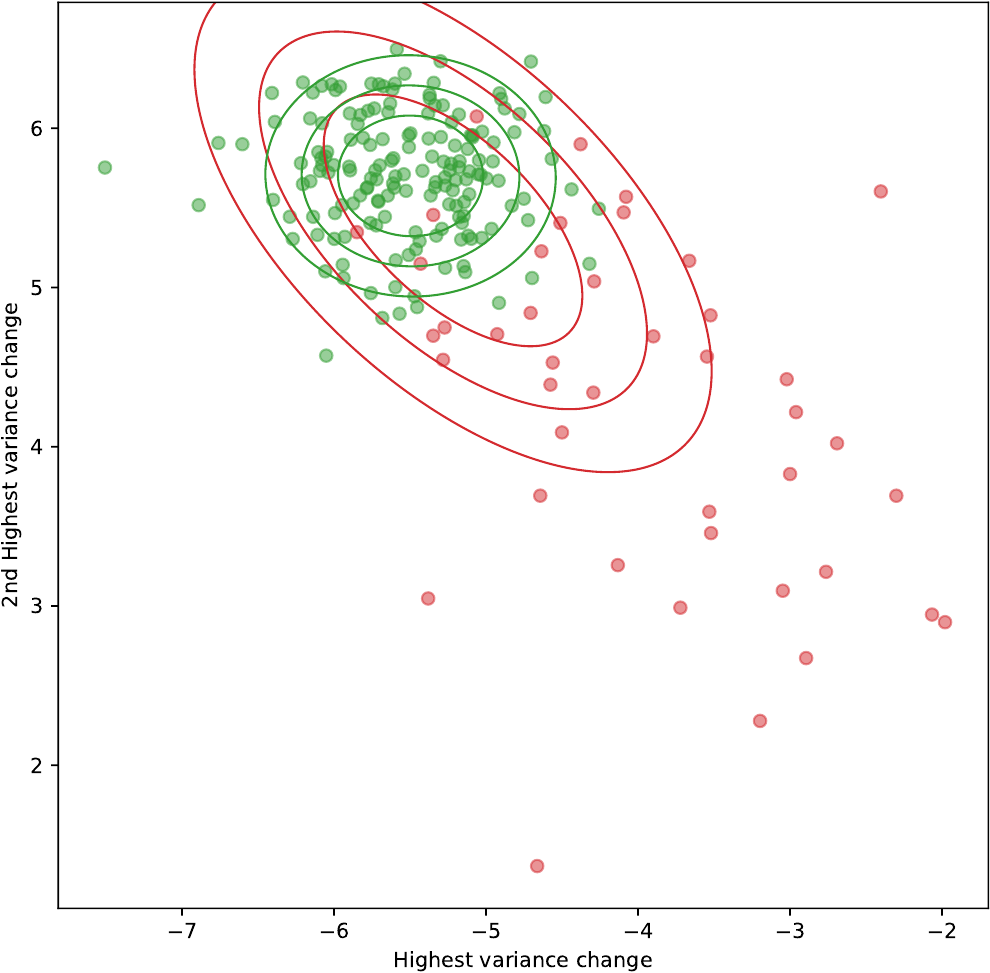}}
\smallskip
\caption{Contours of the fitted multi-variate gaussian (MVG) when the training set contains 0\% (green) and 20\% (red) of pollution for a texture (\textit{carpet}, left) and an object category (\textit{bottle}, right).}
\label{fig:MVG_contours}
\end{figure}

However, as showed in section \ref{subsec:robustness}, there are also discrepancies amongst pre-trained methods. The $k$-Nearest Neighbors approaches (KNN and PatchCore) proved more robust than the Mahalanobis-based ones (Mahalanobis and PaDiM). In order to understand this phenomenon, we pollute two chosen MVTec training sets (\textit{carpet} and \textit{bottle}) with 20\% of defective samples (our maximum pollution ratio), and we extract and process embeddings for these training sets. Then, we fit Mahalanobis on the healthy part of the embedded training set and the full, polluted embedded training set. We finally plot the contours of the learned multi-variate gaussians (MVGs). We plot these contours amongst the axis with the greatest variance change between the healthy and the polluted MVG. Figure \ref{fig:MVG_contours} shows these contours for the two aforementioned categories.

Along the chosen axis, the MVG shifts significantly, thus giving an indication of why Mahalanobis may not be the most robust to pollution. Some defective samples are now much closer to the center of the MVG, and may thus be incorrectly classified as healthy. These plots also give some intuition for the robustness of the two nearest-neighbor approaches (KNN and PatchCore). In these two dimensions, the healthy embeddings appear very close to each other, while the defective embeddings are much more sparse. 

To confirm this intuition, we calculate the average euclidean distance in the embedding space from a polluted sample to another, from a polluted sample to a healthy one, and from a healthy sample to another. We report these distances for \textit{carpet} and \textit{bottle} in Table \ref{tab:distances_table}. We observe that defective samples are almost as far from each other as they are from healthy samples in the embedding space. This explains why the nearest neighbors approaches are so robust to pollution. If all defective samples were close to each other, then many would be classified as healthy, because the distance is used as the anomaly score.

  \begin{table}[ht!]
    \caption{\label{tab:distances_table}Average euclidean distance from healthy to healthy, healthy to defective and defective to defective in the embedding space for the two categories \textit{carpet} and \textit{bottle}. Defective embeddings are almost as far from each other than they they are from healthy embeddings, confirming the intuition on why nearest neighbors approaches are more robust to pollution.} 
    \smallskip
    \begin{tabularx}{\textwidth}{l*{5}{Y}}
     \toprule
     &\multicolumn{2}{c}{\textbf{carpet}}
     &\multicolumn{2}{c}{\textbf{bottle}}\\
     \cmidrule(lr){2-3}\cmidrule(lr){4-5}
     & Healthy & Defective
     & Healthy & Defective\\
     Healthy & 14.66 & 24.07 & 11.92 & 22.66\\
    Defective & 24.07 & 24.61 & 22.66 & 22.90\\
    \bottomrule
    \end{tabularx}
   \end{table}
   
\pagebreak

\section{Conclusion}
\label{sec:conclusion}
In this work, we showed that anomaly detection methods leveraging pre-trained feature extractors are quite robust to polluted datasets, notably compared to self-supervised methods such as CutPaste. We introduced a simple strategy to effectively refine the training data by fitting and evaluating the detectors on the polluted embeddings. We also gave some intuition on why KNN-based methods are more robust to pollution than the ones based on the Mahalanobis distance. It is interesting to note that our refinement strategy is theoretically agnostic of the chosen anomaly detection method, and could thus be used with any. Note that it is also possible to use SROC in combination with a method specifically chosen for refinement (e.g. Mahalanobis), and to use a different method once refinement is performed.

To go further, it would be interesting to make use of the samples predicted as defective, instead of simply removing them. Taking inspiration from the literature\cite{baur2021autoencoders, bergmann2018improving}, one could make use of an generative network (such as an autoencoder) trained with a reconstruction task on the refined data to erase defects from the samples detected as pollution by SROC. This way, defective samples could be transformed into new healthy data in order to improve performance further. Another possibility could be to refine training patches instead of images. This would benefit the pixel-wise methods and limit their drops in performances when refining.

% TO DO (maybe)
% - Mention PatchCore retrieving same images (as a comment for table 2), leading to lower prec/rec
% Sentence on localization results.

% References
\bibliography{report} % bibliography data in report.bib
\bibliographystyle{spiebib} % makes bibtex use spiebib.bst

\pagebreak
\appendix
\section{AU-IoU and AU-PRO curves}
\label{appendix:add-results}

\subsection{Pollution robustness}
\label{appendix:add-results-robust}

Here we provide additional curves to supplement Table \ref{tab:robustness_recap_table}. In particular, we plot the AU-IoU and the AU-PRO as a function of the pollution ratio.
\begin{figure}[hb!]
  \centering
  \includegraphics[width = 300pt]{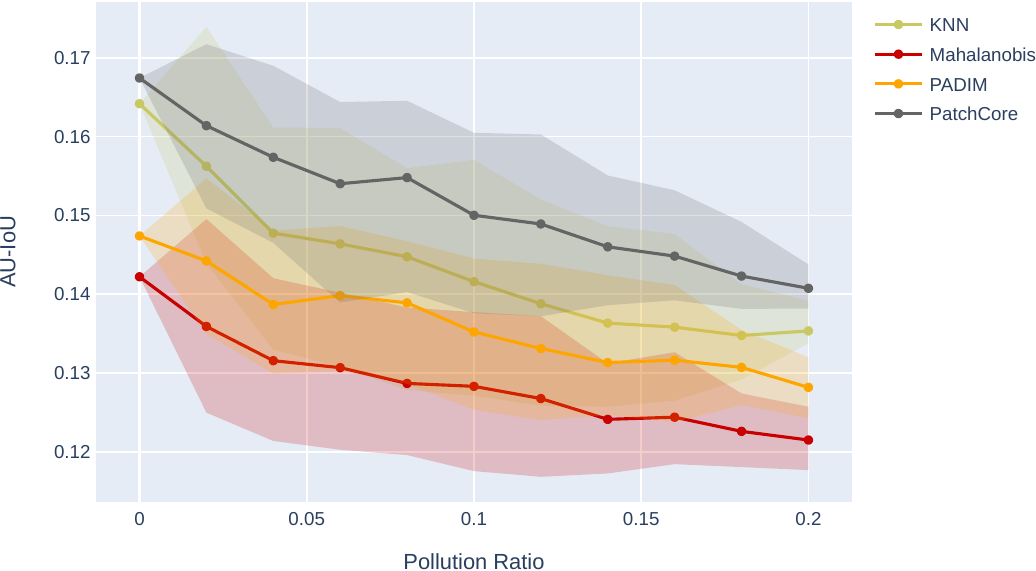}
  \caption{Average AU-IoU across all MVTec AD categories as a function of the pollution ratio, for each detector. Categories \textit{transistor} and \textit{hazelnut} were removed because of their low amount of available defective samples.}
  \label{fig:robustness-iou-curve}
\end{figure}

\begin{figure}[hb!]
  \centering
  \includegraphics[width = 300pt]{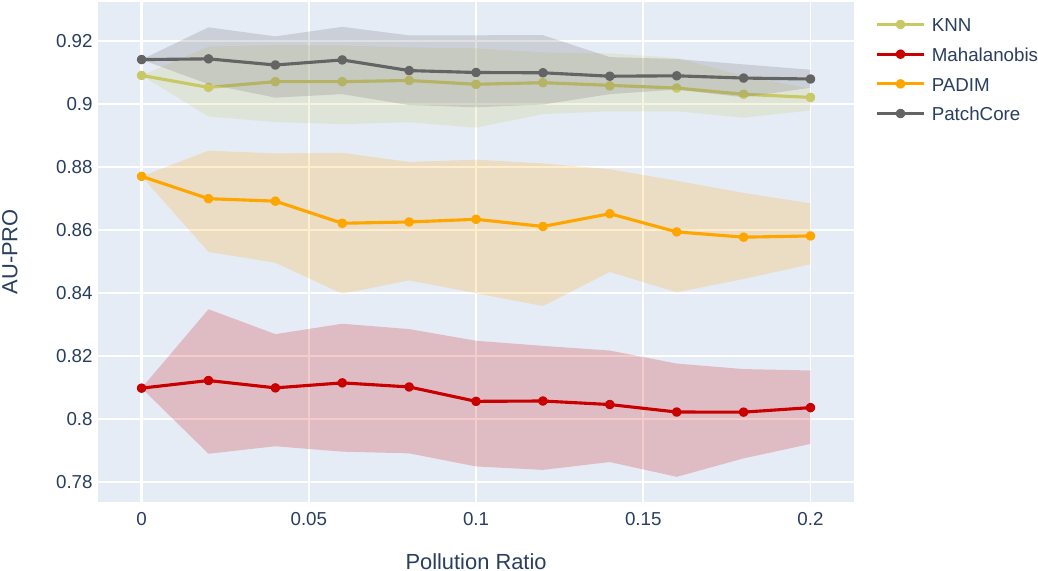}
  \caption{Average AU-PRO across all MVTec AD categories as a function of the pollution ratio, for each detector. Categories \textit{transistor} and \textit{hazelnut} were removed because of their low amount of available defective samples.}
  \label{fig:robustness-pro-curve}
\end{figure}

\pagebreak

\subsection{SROC}
\label{appendix:add-results-sroc}

\begin{figure}[h!]
  \centering
  \includegraphics[width = 300pt]{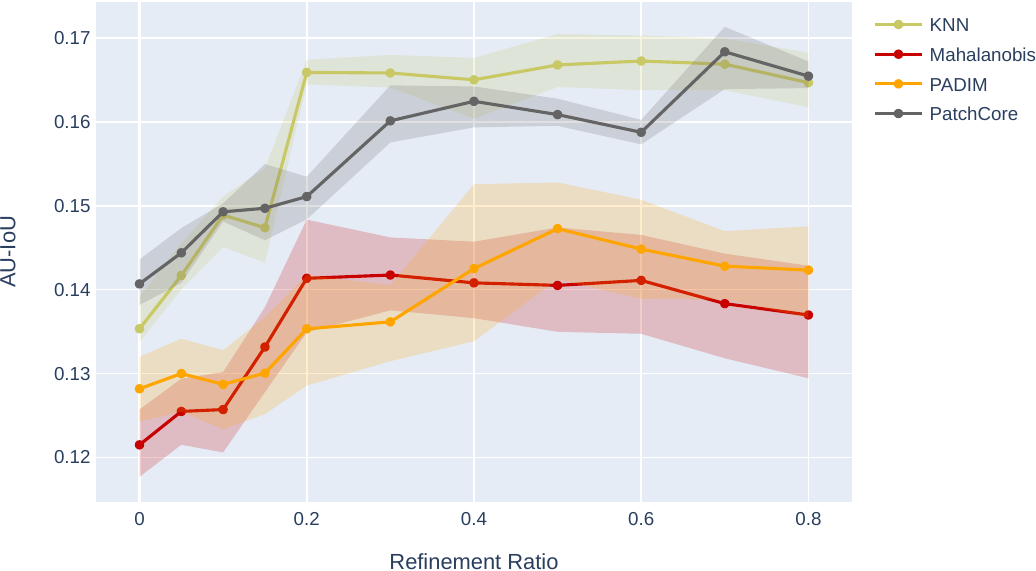}
  \caption{Average AU-IoU across all MVTec AD categories as a function of the refinement ratio, for training sets polluted at a 20\% ratio. Categories \textit{transistor} and \textit{hazelnut} were removed because of their low amount of available defective samples.}
  \label{fig:sroc-iou-curve}
\end{figure}

\begin{figure}[h!]
  \centering
  \includegraphics[width = 300pt]{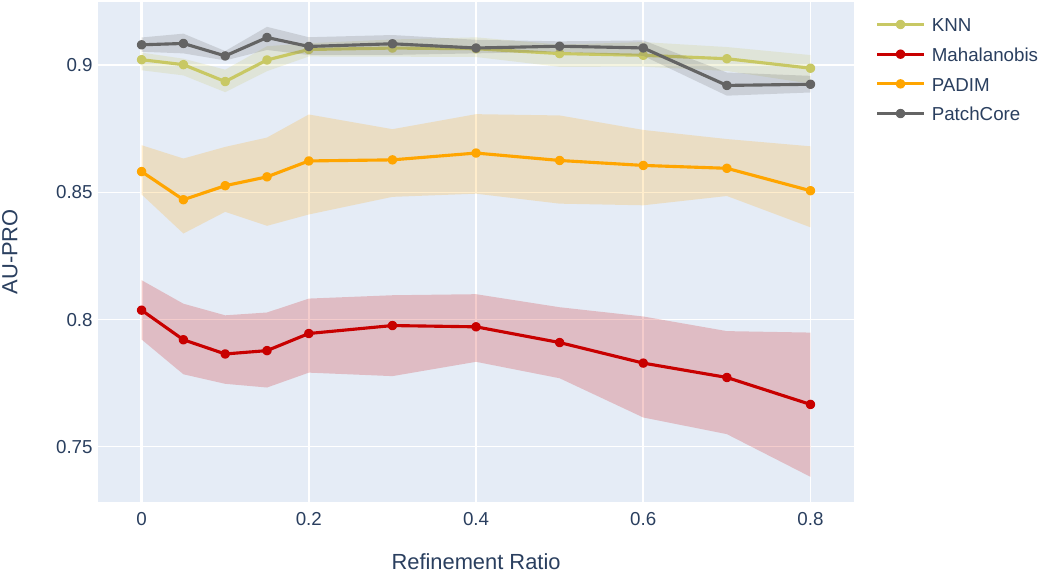}
  \caption{Average AU-PRO across all MVTec AD categories as a function of the refinement ratio, for training sets polluted at a 20\% ratio. Categories \textit{transistor} and \textit{hazelnut} were removed because of their low amount of available defective samples.}
  \label{fig:sroc-pro-curve}
\end{figure}

Figure \ref{fig:sroc-iou-curve} and \ref{fig:sroc-pro-curve} complete Table \ref{tab:naive_recap_table} with AU-IoU and the AU-PRO as a function of the refinement ratio.

\pagebreak
\section{Refinement with Mahalanobis}
\label{appendix:maha-refine}

Finally, we showed in section \ref{subsec:refinement} that amongst the 4 detectors, applying SROC with Mahalanobis gave the best results. Mahalanobis indeed gets the best precision, recall and F1 scores, meaning that more defectives samples are removed (and thus more healthy samples are kept). In this section, we therefore propose to refine the embeddings with Mahalanobis (i.e. we fit and evaluate Mahalanobis on the polluted embeddings and remove the highest scores) before training the other detectors with that refined training data. 
Table \ref{tab:maha_purif_table} shows the results. The main difference is for PaDiM, for which the AUC increases by 1.4 points. This is probably due to the fact that SROC+PaDiM had the lowest precision and recall scores (see Table \ref{tab:naive_recap_table}). Now that we remove more defective samples, we are able to gain more performance back.

  \begin{table}[hb!]
    \centering
    \caption{\label{tab:maha_purif_table}Comparison of SROC when refinement is done with the same detector (\textit{reference}), and when the refinement is done with the Mahalanobis detector (\textit{Mahalanobis}), for training sets polluted at a 20\% ratio.}
    \smallskip
    \begin{tabularx}{\textwidth}{l*{13}{Y} }
     \toprule
     &\multicolumn{3}{c}{KNN}
     &\multicolumn{3}{c}{Mahalanobis}
     &\multicolumn{3}{c}{PaDiM}
     &\multicolumn{3}{c}{PatchCore} \\
     \cmidrule(lr){2-4}\cmidrule(lr){5-7}\cmidrule(lr){8-10}\cmidrule(lr){11-13}
     %\hline
     Refinement & 0\% & 20\% & 40\% & 0\% & 20\% & 40\% & 0\% & 20\% & 40\% & 0\% & 20\% & 40\%\\
     %\hline
     \midrule
     Mean AUC (reference)
     & 88.1 & \textbf{88.9} & 87.4 & 91.8 & \textbf{94.4} & 94.1 & 93.8 & 93.9 & 93.3 & 96.4 & 96.5 & 96.4\\
     %\hline
     Mean AUC (Mahalanobis)
     & 88.1 & 88.6 & 87.5 & 91.8 & \textbf{94.4} & 94.1 & 93.8 & \textbf{95.3} & 94.0 & 96.4 & \textbf{96.8} & 96.6\\
     % \hhline{|=|=|=|=|=|=|=|=|=|=|=|=|=|}$$
     \midrule
     Mean AU-IoU (reference)
     & 0.135 & 0.166 & 0.165 & 0.121 & 0.141 & 0.141 & 0.128 & 0.135 & 0.143 & 0.141 & 0.151 & 0.162\\
     %\hline
     Mean AU-IoU (Mahalanobis)
     & 0.135 & 0.167 & 0.166 & 0.121 & 0.141 & 0.141 & 0.128 & 0.149 & 0.149 & 0.141 & 0.165 & 0.165\\
     %\hhline{|=|=|=|=|=|=|=|=|=|=|=|=|=|}
     \midrule
     Mean AU-PRO (reference)
     & 0.902 & 0.906 & 0.906 & 0.804 & 0.794 & 0.797 & 0.858 & 0.862 & 0.865 & 0.908 & 0.907 & 0.907\\
     %\hline
     Mean AU-PRO (Mahalanobis)
     & 0.902 & 0.904 & 0.903 & 0.804 & 0.794 & 0.797 & 0.858 & 0.868 & 0.873 & 0.908 & 0.914 & 0.913\\
     %\hline
     \bottomrule
    \end{tabularx}
  \end{table}

\begin{figure}[hb!]
  \centering
  \includegraphics[width = 300pt]{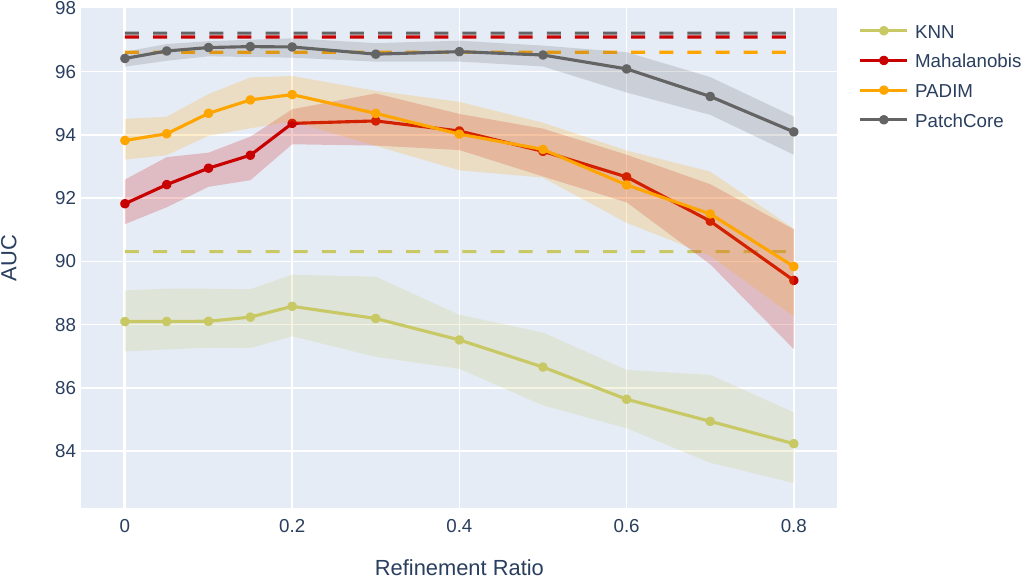}
  \caption{Average AUC of SROC+Mahalanobis across all MVTec AD categories as a function of the refinement ratio, for training sets polluted at a 20\% ratio. Categories \textit{transistor} and \textit{hazelnut} were removed because of their low amount of available defective samples.}
  \label{fig:sroc-AUC-maha}
\end{figure}

\begin{figure}[hb!]
  \centering
  \includegraphics[width = 300pt]{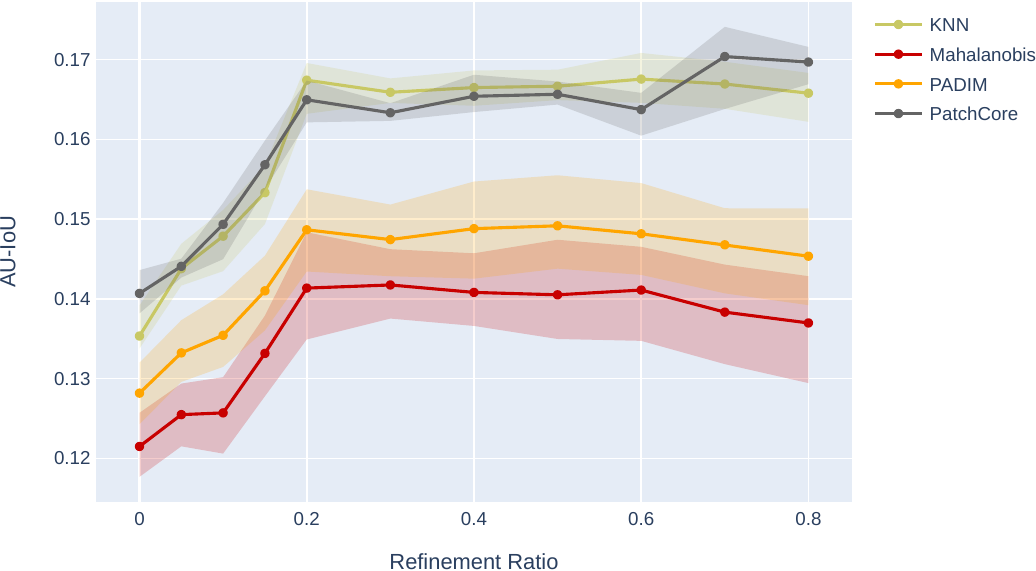}
  \caption{Average AU-IoU of SROC+Mahalanobis across all MVTec AD categories as a function of the refinement ratio, for training sets polluted at a 20\% ratio. Categories \textit{transistor} and \textit{hazelnut} were removed because of their low amount of available defective samples.}
  \label{fig:sroc-iou-maha}
\end{figure}

\begin{figure}[hb!]
  \centering
  \includegraphics[width = 300pt]{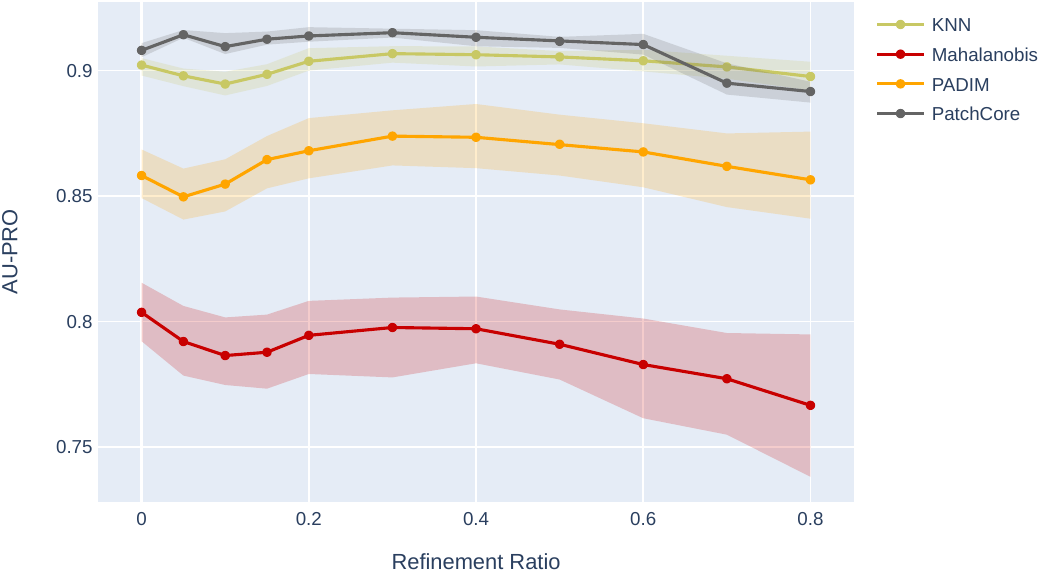}
  \caption{Average AU-PRO of SROC+Mahalanobis across all MVTec AD categories as a function of the refinement ratio, for training sets polluted at a 20\% ratio. Categories \textit{transistor} and \textit{hazelnut} were removed because of their low amount of available defective samples.}
  \label{fig:sroc-pro-maha}
\end{figure}

\end{document}